\documentclass{article}

\usepackage[final,nonatbib]{neurips_2020}

\usepackage[utf8]{inputenc} 
\usepackage[T1]{fontenc}    
\usepackage[colorlinks,citecolor=teal]{hyperref}   
\usepackage{url}            
\usepackage{booktabs}       
\usepackage{amsfonts}       
\usepackage{nicefrac}       
\usepackage{microtype}      
\usepackage{graphicx}
\usepackage{comment}
\usepackage{dirtytalk}
\usepackage{amsmath}
\usepackage{xcolor}
\usepackage{booktabs}
\usepackage{arydshln}
\usepackage[caption=false]{subfig}
\usepackage{enumitem}	      

\def\eg{\emph{e.g.,}}           
\def\ie{\emph{i.e.,}}           
\def\vs{\emph{vs.}}                                
\def\etal{\emph{et al.}}                           

\title{%
   Self-Supervised Visual Representation Learning\\from Hierarchical Grouping%
}

\author{%
  Xiao Zhang\\
  University of Chicago\\
  \texttt{zhang7@uchicago.edu}
  \And
  Michael Maire\\
  University of Chicago\\
  \texttt{mmaire@uchicago.edu}%
}

\begin{document}
\maketitle

\begin{abstract}
We create a framework for bootstrapping visual representation learning from a primitive visual grouping capability. We operationalize grouping via a contour detector that partitions an image into regions, followed by merging of those regions into a tree hierarchy. A small supervised dataset suffices for training this grouping primitive. Across a large unlabeled dataset, we apply this learned primitive to automatically predict hierarchical region structure. These predictions serve as guidance for self-supervised contrastive feature learning: we task a deep network with producing per-pixel embeddings whose pairwise distances respect the region hierarchy. Experiments demonstrate that our approach can serve as state-of-the-art generic pre-training, benefiting downstream tasks. We additionally explore applications to semantic region search and video-based object instance tracking.

\end{abstract}

\section{Introduction}
\label{sec:intro}

The ability to learn from large unlabeled datasets appears crucial to deploying
machine learning techniques across many application domains for which
annotating data is too costly or simply infeasible due to scale.  For visual
data, quantity and collection rate often far outpace ability to annotate,
making self-supervised approaches particularly crucial to future advancement of
the field.

Recent efforts on self-supervised visual learning fall into several broad
camps.  Among them, Kingma~\etal~\cite{kingma2013auto} and
Donahue~\etal~\cite{donahue2016adversarial} design general architectures to
learn latent feature representations, driven by modeling image distributions.
Another group of approaches~\cite{gidaris2018unsupervised,
doersch2015unsupervised,larsson2016learning,noroozi2016unsupervised} leverage,
as supervision, pseudo-labels automatically generated from hand-designed proxy
tasks.  Here, the general strategy is to split data examples into two parts and
predict one from the other.  Alternatively, Wu~\etal~\cite{wu2018unsupervised}
and Zhuang~\etal~\cite{zhuang2019local} learn visual features by asking a deep
network to embed the entire training dataset, mapping each image to a location
different from others, and relying on this constraint to drive the emergence of
a topology that reflects semantics.  Across many of these camps, technical
improvements to the scale and efficiency of learning further boost
results~\cite{he2019momentum,chen2020simple,oord2018representation,
donahue2019large}.  Section~\ref{sec:related} provides a more complete
background.

\begin{figure}[!tp]
   \centering
   \includegraphics[width=\textwidth]{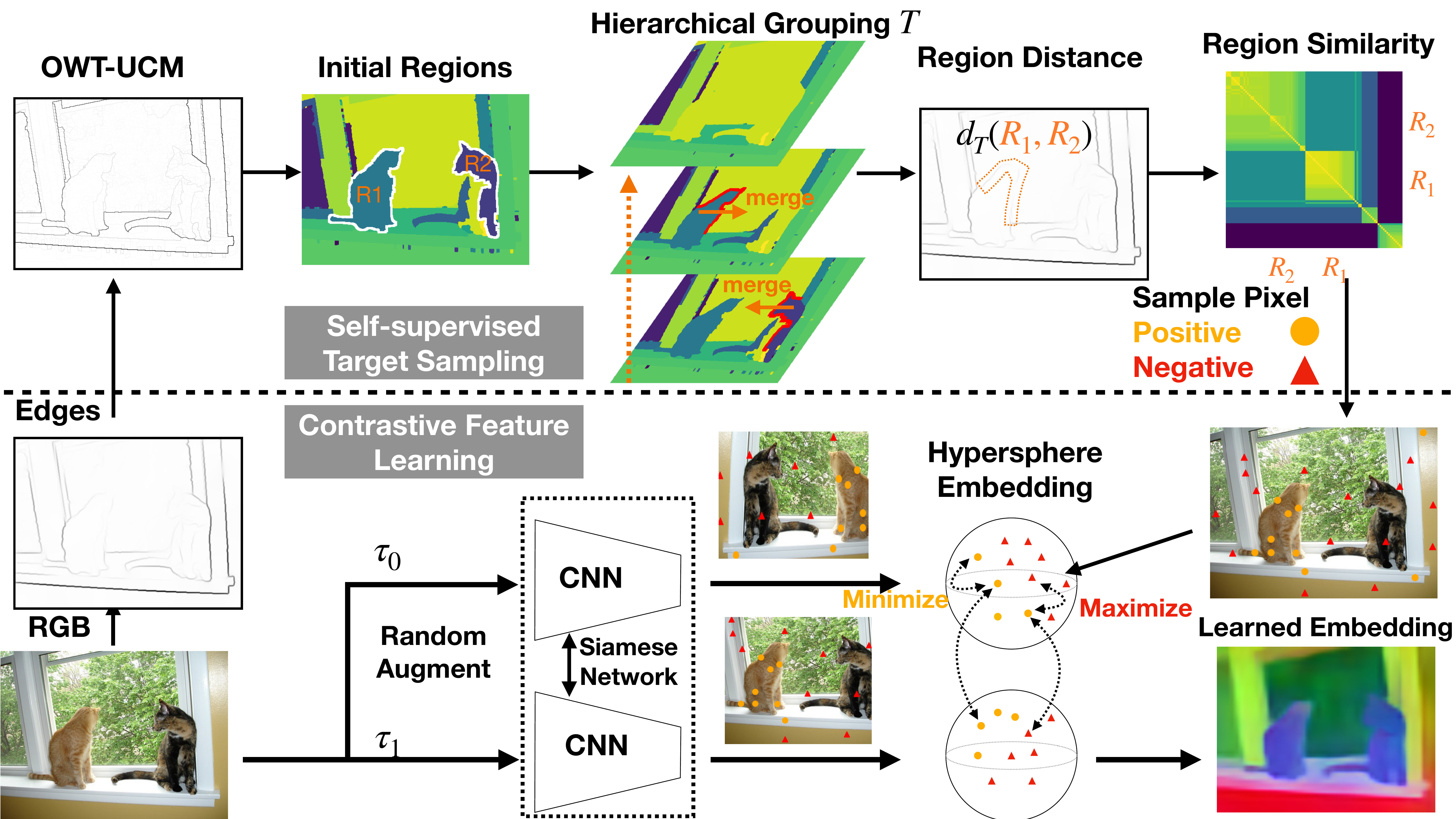}
   \caption{%
      \textbf{Bootstrapping semantic representation learning via primitive
      hierarchical grouping.}
      \textbf{\emph{Top: Self-Supervised Target Sampling.}}
      From a hierarchical segmentation of an image (\ie~a region tree $T$),
      rendered here as a boundary strength map
      (OWT-UCM~\cite{arbelaez2010contour}), we define distance between regions
      $d_T(R_1,R_2)$ according to the level in the hierarchy at which they
      merge.  Treating this distance as a similarity measure between pixels, we
      sample \textcolor{orange}{positive} and \textcolor{red}{negative} pairs
      of pixels, according to their grouping likelihood in the hierarchy.
      \textbf{\emph{Bottom: Contrastive Feature Learning.}}
      A contour detector~\cite{arbelaez2010contour}, trained on a small
      dataset~\cite{BSDS}, produces hierarchical segmentations across a larger
      unlabeled image set.  Automatically extracted positive and negative pixel
      pairs serve to drive a from-scratch initialized CNN to learn to predict
      pixel-wise embeddings which respect the region hierarchy.  Unlike
      SegSort~\cite{hwang2019segsort}, our pipeline does not merely fine-tune
      ImageNet~\cite{deng2009imagenet} pre-trained models for semantic
      segmentation, but instead addresses representation learning entirely
      from scratch.%
   }
   \label{fig:iconic}
\end{figure}

\begin{figure}[t]
   \centering
   \includegraphics[width=\textwidth]{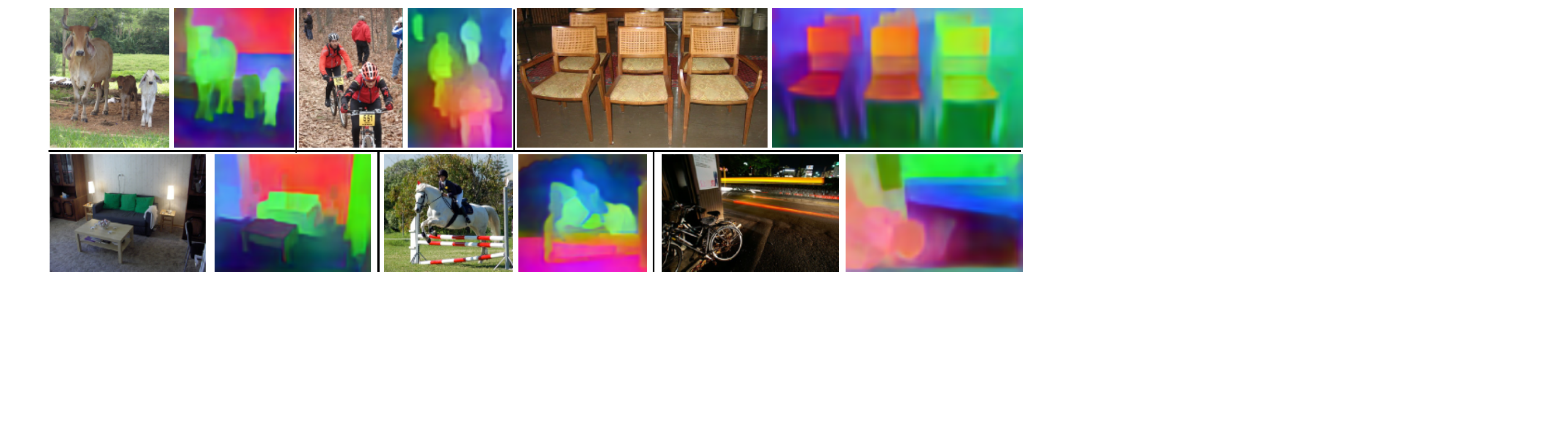}
   \caption{%
      \textbf{Visualization of feature embeddings.}
      We apply PCA to the embeddings produced by a CNN trained using the
      self-supervised bootstrapping approach of Figure~\ref{fig:iconic}.  On
      validation examples, we display the first three PCA components as an RGB
      image.  The output feature representations capture aspects of semantic
      category (\eg~top left) and object instance identity (\eg~top right).
   }
   \label{fig:iconic_app}
   \vspace{-10pt}
\end{figure}

We approach self-supervised learning using a strategy somewhat different from
those approaches outlined above.  While incorporating aspects of proxy tasks
and embedding objectives, a key distinction is that our system's proxy task is
itself generated by a (simpler) trained vision system.  We thus seek to
bootstrap visual representation learning in a manner loosely inspired by,
though certainly not accurately mirroring, a progression of simple to complex
biological vision systems.  This is an under-explored, though not unrecognized,
strategy in computer vision.  Serving as a noteworthy example is
Li~\etal's~\cite{li2016unsupervised} approach of using motion as a readily
available, automatically-derived, supervisory signal for learning to detect
contours in static images.   We focus on the next logical stage in such a
bootstrapping sequence: using a pre-existing contour detector to automatically
define a task objective for learning semantic visual representations.
Figure~\ref{fig:iconic} illustrates how this primitive visual system,
combined with a modern contrastive feature learning framework, trains a
convolutional neural network (CNN) to produce semantic embeddings
(Figure~\ref{fig:iconic_app}).  We defer full details to
Section~\ref{sec:method}.

Our system not only leverages contours to learn visual semantics, but also
leverages a small amount of annotated data to learn from a vastly larger pool
of unlabeled data.  Our visual primitive of contour detection is trained in a
supervised manner from only $500$ annotated images~\cite{BSDS}.  This primitive
then drives self-supervised learning across datasets ranging from tens of
thousands to millions of images; in this latter phase, our system does not
utilize any annotations and trains from randomly initialized parameters.
This is a crucial distinction from SegSort~\cite{hwang2019segsort}, whose
pipeline bears coarse resemblance to our Figure~\ref{fig:iconic}.  SegSort's
``unsupervised'' learning setting still relies on starting from ImageNet
pre-trained CNNs; its ``unsupervised'' aspect is only with respect to forgoing
use of segmentation ground-truth.  In contrast, we address the problem of
representation learning entirely from scratch, save for the $500$ annotated
images of the Berkeley Segmentation Dataset~\cite{BSDS}.

ImageNet, even without labels, is curated: most ImageNet images contain a
single category.  This provides some implicit supervision, which may bias the
self-supervised work that experimentally targets learning from ImageNet,
including MoCo~\cite{he2019momentum}, InstFeat~\cite{ye2019unsupervised} and
others~\cite{wu2018unsupervised,chen2020simple}.  Many use cases for
self-supervision will lack such curation.  As our bootstrapping strategy
utilizes a visual primitive geared toward partitioning complex scenes into
meaningful components, it is a better fit to learning on unlabeled examples
from datasets containing scenes (\eg~PASCAL~\cite{everingham2010pascal},
COCO~\cite{lin2014microsoft}).

Using a similar siamese network, we outperform
InstFeat~\cite{ye2019unsupervised} by a large margin on the task of learning
transferable representations from PASCAL and COCO images alone (disregarding
labels).  In this setting, our results are competitive with those of the
state-of-the-art MoCo system~\cite{he2019momentum}, while our method remains
simpler.  We do not rely on a momentum encoder or memory bank.  Even with this
simpler training architecture, our segmentation-aware approach enhances the
efficiency of learning from complex scenes.  Here, our pre-training converges
in under half the epochs needed by MoCo to learn representations with
comparable transfer performance on downstream tasks.

In addition to evaluating learned representation quality on standard
classification tasks, Sections~\ref{sec:experiments} and~\ref{sec:results}
explore applications to semantic region search and instance tracking in
video.  Using similarity in our learned feature space to conduct matching
across images and frames, we outperform competing methods in both applications.
Our results point to a promising new pathway of crafting self-supervised
learning strategies around bootstrapping the training of one visual module
from another.

\section{Related Work}
\label{sec:related}

\textbf{Self-Supervised Representation Learning.}
Approaches to self-supervised visual learning that train networks to predict
one aspect of the data from another have utilized a variety of proxy tasks,
including prediction of
context~\cite{pathakCVPR16context},
colorization~\cite{zhang2016colorful,larsson2017colorization},
cross-channels~\cite{zhang2017split},
optical-flow~\cite{zhan2019self}, and
rotation~\cite{gidaris2018unsupervised}.
Oord~\etal~\cite{oord2018representation} train to predict future
representations in the latent space of an autoregressive model.
As objects move coherently in video, Mahendran~\etal~\cite{mahendran2018cross}
learn pixel-wise embeddings for static frames, such that their pair-wise
similarity mirrors that of optical flow vectors.

Another family of methods casts representation learning in terms of clustering
or embedding objectives.  DeepCluster~\cite{caron2018deep} iterates between
clustering CNN output representations to define target classes and re-training
the CNN to better predict those targets.  Contrastive multiview
coding~\cite{tian2019contrastive} sets the objective as mapping different
views of the same scene to a common embedding location, distinct from other
scenes.  Ye~\etal~\cite{ye2019unsupervised} apply similar intuition with
respect to data augmentation.  Instance discrimination~\cite{
wu2018unsupervised} formulates feature learning as a non-parametric softmax
prediction problem, enforcing consistency between a predicted hypersphere
embedding and a counterpart maintained in memory banks.  Following
Wu~\etal~\cite{wu2018unsupervised}, Zhuang~\etal's local aggregation
approach~\cite{zhuang2019local} uses additional clustering steps to reason
about embedding targets.

Differing from the siamese network of Ye~\etal~\cite{ye2019unsupervised} and
the dataset-level memory bank of Wu~\etal~\cite{wu2018unsupervised},
momentum contrast (MoCo)~\cite{he2019momentum} uses a moving-average
encoder for reference embeddings.  This offers scalability superior to a
memory bank.  Self-supervised networks trained with MoCo outperform
their ImageNet-supervised counterparts, as benchmarked by fine-tuning to
multiple tasks.

Like MoCo, our approach also benefits from a feature learning paradigm that
jointly considers augmentation invariance and negative example sampling.
But, instead of learning feature representation only at image level, our method
learns pixel-wise semantic affinity in the context of regions.  Our system
relies on a simpler siamese architecture, rather than requiring a
moving-average encoder.

\textbf{Image Segmentation.}
The classic notion of image segmentation -- partitioning into meaningful
regions without necessarily labeling according to known semantic classes --
has a rich history.  Given the duality between regions and their bounding
contours, modern approaches often focus on the problem of contour detection.
Though deep neural networks are now the dominant tool for contour
detection~\cite{shen2015deepcontour,bertasius2015high,xie2015holistically,
kokkinos2015pushing,Yang_2016_CVPR}, the best prior approaches~\cite{
arbelaez2010contour,arbelaez2014multiscale,dollar2014fast} deliver somewhat
respectable results.

Pre-training CNNs on ImageNet before fine-tuning them for contour detection
provides accuracy gains~\cite{xie2015holistically}.  But, as our purpose
is to bootstrap representation learning from contours, obviating the need for
ImageNet supervision, we do not want ImageNet labels used in our contour
detector training pipeline.  For experiments, we therefore select an older
detector, based on random forests~\cite{dollar2014fast}, along with
traditional machinery (OWT-UCM)~\cite{arbelaez2010contour} for converting
contours into region hierarchies.  Both components are trained using only
the $500$ labeled images of the BSDS~\cite{BSDS}.

\textbf{Representation Learning using Segmentation.}
Several works incorporate segmentation into representation learning.
Fathi~\etal~\cite{fathi2017semantic} formulate the task of instance
segmentation as metric learning.  They train a network for instance-aware
embedding by optimizing feature similarity of pixels sampled within or
cross-instances.  Kong~\etal~\cite{kong2018recurrent}, adopting a similar
objective, combine a CNN with a differentiable mean-shift clustering approach
to learn instance segmentation.  Chen~\etal~\cite{chen2018blazingly} address
video instance segmentation via pixel embedding with a modified triplet loss.
Pathak~\etal~\cite{pathak2017learning} learn representations for recognition
tasks by predict moving object segmentation from static frames.

SegSort~\cite{hwang2019segsort} utilizes an iterative grouping algorithm in
EM (expectation maximization) fashion, learning a segmentation-aware embedding
of pixels onto a hypersphere.  Specifically, it leverages regions
(computed by OWT-UCM~\cite{arbelaez2010contour} on
HED~\cite{xie2015holistically} contours) as defining separation criteria,
maximizing pairwise intra-similarity and inter-contrast for the pixels
within the same or different regions, respectively.  Unlike supervised
counterparts, SegSort can learn semantic segmentation on top of
ImageNet~\cite{deng2009imagenet} pre-trained CNNs.  In this mode, contours
(as opposed to ground-truth per-pixel class labels) provide the only
additional supervisory signal for learning region semantics.

While our work shares a similar spirit with SegSort~\cite{hwang2019segsort},
we aim to bootstrap learning of semantic region representations entirely
from scratch, removing the dependence on ImageNet pre-training for both the
primary CNN and the contour detection component.

\section{Bootstrapping Semantics from Grouping}
\label{sec:method}

We train a convolutional neural network $\phi(\cdot)$, which maps an input
image $\bold{I}$ into a spatially-extended feature representation
$\bold{F} = \phi(\bold{I})$.  Let $\bold{F}(i) \in \mathbb{R}^d$ denote the
output $d$-dimensional feature embedding of pixel $i$.  We adopt a contrastive
learning objective that operates on a pixel level.  Defining
$\text{sim}(\bold{F}(i),\bold{F}(j))
   = \bold{F}(i)^T\bold{F}(j)/(||\bold{F}(i)||\cdot||\bold{F}(j)||)$
as the cosine similarity between feature vectors $\bold{F}(i)$ and
$\bold{F}(j)$, we want to learn optimal network parameters $\phi^*$ as
follows:
\begin{eqnarray}
   \phi^* = \underset{\phi}{\operatorname{arg\,max}} \sum_{i}\sum_{m\in Pos(i)}
      \frac{\exp(\text{sim}(\bold{F}(i),\bold{F}(m)))}%
           {\exp(\text{sim}(\bold{F}(i),\bold{F}(m))) +
            \sum_{n\in Neg(i)}\exp(\text{sim}(\bold{F}(i),\bold{F}(n)))}
   \label{eqn:main_eqn}
\end{eqnarray}
where $Pos(i)$, $Neg(i)$ denote the pixels in the same or different
semantic categories as pixel $i$.

Unlike the supervised setting, where annotations determine $Pos(i)$, $Neg(i)$,
we must automate estimation of these relationships.  In designing such a
procedure, we operate under stringent assumptions: (1) the network is
initialized from scratch, and (2) training images may contain complex scenes.

\textbf{Grouping Primitive.}
We deriving guidance from a visual grouping primitive to sample $Pos(i)$ and
$Neg(i)$.  Specifically, we adopt a contour detector $\phi_E$, which, acting
on image $I$, produces an edge strength map $E = \phi_{E}(I)$.  We then
convert $E$ into a region map $R$ using a hierarchical merging process which
repeatedly removes the weakest edge separating two regions.  The real-valued
edge strength at which two distinct regions merge defines a distance
metric, which we extend to a notion of distance between pixels.  This is
precisely the procedure for constructing an ultrametric contour map
(UCM)~\cite{arbelaez2006boundary}, which can equivalently be regarded as
as both a reweighted edge map, and a tree $T$ defining a region hierarchy.
The leaves of $T$ are the initial, finest-scale, regions; interior nodes
represent larger regions formed by merging their children and have a
real-valued height in the hierarchy equal to the distance between their
child regions.  Figure~\ref{fig:iconic} (top, center) shows an example.

As remarked upon earlier, we utilize the structured forest edge detector of
Doll{\'a}r and Zitnick~\cite{dollar2014fast}, trained on a small dataset
(BSDS~\cite{BSDS}).  In constructing the region tree, we follow
Arbel{\'a}ez~\etal~\cite{arbelaez2010contour}, applying their variant of the
oriented watershed transform (OWT) prior to computing the UCM.

\textbf{Pixel Sampling from Region Metric.}
Denote the distance between regions $R_1$ and $R_2$ defined by the UCM-derived
region tree $T$ as $d_T(R_1,R_2)$.  We lift this region distance to serve as
a measure of the probability that two pixels belong in the same semantic
region.  Considering two pixels $i,j$, we express $P(j\in Pos(i))$, the
probability to assign the pixel $j$ into the positive set of $i$, as a function
of region distance.  For $i \in R_a, j\in R_b$:
\begin{eqnarray}
   P(j\in Pos(i))
      = \frac{\exp(- d_T(R_a, R_b)/\sigma_p)}%
             {\sum_{m\neq a}^M\exp(-d_T(R_a, R_m)/\sigma_p) + 1}
\end{eqnarray}
where $\sigma_p$ is a temperature parameter to control the concentration on
region distance, and we manually cast the self-similarity to one:
$\exp(- d_T(R_a, R_a)/\sigma_p)\rightarrow 1$
to further concentrate positive sampling within the anchor regions.
In experiments, we find this trick leads to better positive sampling and
performance.  Similarly, we can define the probability of assigning $j$ to
$i$'s negative set as:
\begin{eqnarray}
   P(j\in Neg(i))
      = \frac{\exp(d_T(R_a, R_b)/\sigma_p)}
             {\sum_{m\neq a}^M\exp(d_T(R_a, R_m)/\sigma_p)}
\end{eqnarray}
Note that we do not formulate $P(j\in Neg(i))$ as the complement of
$P(j\in Pos(i))$, for the purpose of excluding the pixels whose assignments
are vague.  With these defined distributions, we can sample corresponding
reference pixels from the image.

\textbf{Augmented Reference Sampling.}
Asking vision systems to produce feature representations that are invariant to
common image transformations (\eg~color change, object deformation,
occlusion) appears to be extremely helpful in learning semantic features.  To
this end, we augment our sampling of positive and negative pixel pairs in
two ways:
\vspace{-5pt}
\begin{itemize}[leftmargin=0.15in]
\item{%
   \textit{\textbf{Image Augmentation.}}
   We impose data augmentation on training images to collect extra positive
   and negative pairs.  Specifically, suppose we have sampled $Pos(i)$ and
   $Neg(i)$ for pixel $i$, and augmented input image with transformation
   $\tau$.  We can then have augmented pixels pairs
   $Pos(i)_{aug}=\tau(Pos(i))$ and $Neg(i)_{aug} = \tau(Neg(i)))$
   to sample referenced features on $\phi(\tau(\bold{I}))$.%
}
\item{%
   \textit{\textbf{Crossing-image Negative Sampling ($cns$).}}
   Capturing object-level, rather than merely instance-level, semantics
   requires considering relationships across the entire dataset; we need a
   signal relating reference pixels from different images.  We adopt an
   approach similar
   to~\cite{wu2018unsupervised,he2019momentum,wang2019learning} and randomly
   sample negative features from the dataset (but operationalized on a
   per-pixel level).  We denote these randomly sampled negatives as
   $Neg(i)_{cns}$ for selected anchor pixel $i$.%
}
\end{itemize}

\textbf{Optimization.}
By jointly considering all sources to provide reference feature vectors,
we can modify our target in Equation~(\ref{eqn:main_eqn}) by updating:
$Pos(i)$ with $Pos^{+}(i) = Pos(i) \cup Pos(i)_{aug}$ and
$Neg(i)$ with $Neg^{+}(i) = Neg(i) \cup Neg(i)_{aug} \cup Neg(i)_{cns}$.
During training, we end-to-end optimize Equation~(\ref{eqn:main_eqn}) starting
from a randomly initialized network $\phi$, as to produce feature vectors
$\bold{F}(i)$ which encode fine-grained pixel-wise semantic representations.

\section{Experiment Settings}
\label{sec:experiments}

\subsection{Datasets and Preprocessing}

\textbf{Datasets.}
We experiment on datasets of complex scenes, with variable numbers of object
instances: PASCAL~\cite{everingham2010pascal} and
COCO~\cite{lin2014microsoft}.  PASCAL provides 1464 and 1449 pixel-wise
annotated images for training and validation, respectively.
Hariharan~\etal~\cite{hariharan2011semantic} extend this set with extra
annotations, yielding 10582 training images, denoted as $\textit{train\_aug}$.
The COCO-2014~\cite{lin2014microsoft} dataset provides instance and semantic
segmentations for 81 foreground object classes on over 80K training images.
In unsupervised experiments, we train the network on the union of images in the
$\textit{train\_aug}$ set of PASCAL and $\textit{train}$ set of COCO.  We
evaluate learned embeddings on the PASCAL $\textit{val}$ set by training a
pixel-wise classifier for semantic segmentation on PASCAL $\textit{train\_aug}$,
set atop frozen features.

We also benchmark learned embeddings on the
DAVIS-2017~\cite{Pont-Tuset_arXiv_2017} dataset for the task of instance mask
tracking.  We $\textit{ONLY}$ train on PASCAL $\textit{train\_aug}$ and COCO
$\textit{train}$, without including any images from DAVIS-2017.  After
self-supervised training, we directly evaluate on the $\textit{val}$ set of
DAVIS-2017 as to propagate initial instance masks to consecutive frames through
embedding similarity.

\textbf{Edges and Regions.}
Our proposed self-supervised learning framework starts with an edge predictor.
We avoid neural network based edge predictors like
HED~\cite{xie2015holistically}, which depend upon an ImageNet pre-trained
backbone.  We instead turn to structured edges (SE)~\cite{dollar2014fast},
which only leverages the small supervised BSDS~\cite{BSDS} for training.
From SE-produced edges, we additionally compute a spectral edge signal
(following~\cite{arbelaez2010contour}), and feed a combination of both edge
maps into the processing steps of OWT-UCM~\cite{arbelaez2010contour}.  This
converts the predicted edges a region tree $T$, which we constrain to have
no more than $40$ regions at its finest level.

\subsection{Implementation Details}
\textbf{Network Design.}
We use a randomly initialized ResNet-50~\cite{he2015deep} backbone.  To
produce high fidelity semantic features, we adopt a hypercolumn
design~\cite{hariharan2015hypercolumns} that combines feature maps coming from
different blocks.  We keep the stride of ResNet-50 unchanged and adopt a
single $3 \times 3$ Conv-BN-ReLU block to project the last outputs of the
Res3 and Res4 blocks into 256-channel feature maps.  These two feature maps
are both interpolated to match the spatial resolution of a $4 \times$
downsampled input image before concatenation.  Finally, we project the
resulting feature map using a single $3 \times 3$ convolution layer, to
produce a final 32-channel feature map as our output semantic embedding
$\bold{F}$.

\textbf{Training.}
We use Adam~\cite{adam} to train our model for 80 epochs with batch size 70.
We initialize learning rate as 1e-2 which is then decayed by 0.1 at 25, 45, 60
epochs, respectively.  We perform data augmentation including random resized
cropping, random horizontal flipping, and color jittering on input images,
which are then resized to $224 \times 224$ before being fed into the network.
For one image, we randomly sample 7 regions and, for each region, sample 10
positive pixels and 5 negative pixels.  We use $\sigma_p = 0.8$ for all
experiments.

In experiments fine-tuning on PASCAL $train\_aug$, we freeze all the base
features and only update parameters of a newly added head, the Atrous Spatial
Pyramid Pooling (ASPP) module from DeepLab V3~\cite{chen2017rethinking}.  Here,
we use SGD with weight decay $5e^{-4}$ and momentum 0.9 to optimize the
pixel-wise cross entropy loss for 20K iterations with batch size 20.  We
randomly crop and resize images to 384x384 patches.  The learning rate starts
at 0.03 and decays by 0.1 at 10K and 15K iterations.

\section{Results and Analysis}
\label{sec:results}

\subsection{Evaluation of Learned Self-Supervised Representations}

To compare with other representation learning approaches:
InstFeat~\cite{ye2019unsupervised},
MoCo~\cite{he2019momentum}, and
SegSort~\cite{hwang2019segsort}, we use the official code released by the
respective authors to train a ResNet-50 backbone on the same unified dataset of
COCO $train$ and PASCAL $train\_aug$ set.  We preserve the default parameters
of these approaches.  For InstFeat~\cite{ye2019unsupervised} and
SegSort~\cite{hwang2019segsort}, we set the total training epochs to 80 and
rescale the timing of learning rate decay accordingly.  We find 80 epochs
is sufficient for convergence of these models.

For SegSort~\cite{hwang2019segsort}, we replace their network, which is built
upon an ImageNet pre-trained ResNet-101, with a ResNet-50 that is initialized
from scratch.  We update the related parameters and region map, \eg~learning
rate, decayed epochs, total epochs, to be consistent with our setting.  This
allows us to compare with SegSort as a truly unsupervised approach, rather
than one that relies on ImageNet pre-training.
Table~\ref{table:finetune_result} reports results after fine-tuning for PASCAL
semantic segmentation.

\begin{table}[!t]
\centering
\begin{tabular}{cccccc}
\toprule
 Method & \begin{tabular}{@{}c@{}}Cross-Image  \\ Negative  Sample \end{tabular} & \begin{tabular}{@{}c@{}}Unsupervised \\ Pre-train \end{tabular} & \begin{tabular}{@{}c@{}}Pre-train \\ Epochs \end{tabular} & mIoU\\
\specialrule{.2em}{.1em}{.1em}
Random Init  & None & None & - & 12.99\\
Random Init(End to end) & None & None & - &  45.43\\
ImageNet Pre-train & None & None & - &  70.54\\
\specialrule{.2em}{.1em}{.1em}
MoCo~\cite{he2019momentum} & Image & ImageNet & 200& 65.53\\
Ours &  Pixel(cns=5) & ImageNet & 200& 51.02 \\
$\textbf{Ours}^\dagger$ & Pixel(cns=5) + Image & ImageNet & 200& 64.70 \\
\specialrule{.2em}{.1em}{.1em}
InstFeat~\cite{ye2019unsupervised} & Image & PASCAL + COCO &80 & 38.11\\
SegSort~\cite{hwang2019segsort} & None & PASCAL + COCO &80 & 36.15\\
MoCo~\cite{he2019momentum} & Image & PASCAL + COCO & 80 & 42.07\\
MoCo~\cite{he2019momentum} & Image & PASCAL + COCO & 200 & 47.01\\
\textbf{Ours} & Pixel(cns=5,20) & PASCAL + COCO& 80 & 46.51\\
\\[-9pt]\hdashline\\[-8pt]
\textbf{Ours}(HED) & Pixel(cns=5,20) & PASCAL + COCO& 80 & 48.82\\
\midrule
MoCo~\cite{he2019momentum} & Image  & PASCAL & 400 & 32.04\\
Ours &Pixel(cns=5)  & PASCAL & 400 & 43.51\\
Ours & Pixel(cns=5)  & PASCAL + COCO $\times$ 0.5 & 140 & 44.92\\
Ours &Pixel(cns=5)  & PASCAL + COCO& 80 & 45.57\\
Ours  & None & PASCAL + COCO& 80 & 39.63\\
\specialrule{.2em}{.1em}{.1em}
\end{tabular}
\vspace{5pt}
\caption{%
   \textbf{Quantitative evaluation of learned self-supervised representations.}
   We train classifiers on top of frozen features to predict semantic
   segmentation, and benchmark on the PASCAL $val$ set.  When using complex
   images (PASCAL + COCO) for unsupervised pre-training \emph{(bottom half of
   table)}, our approach outperforms~\cite{ye2019unsupervised,
   hwang2019segsort}.  Moreover, without using a momentum encoder or memory
   bank, we exceed MoCo~\cite{he2019momentum} on training efficiency (better
   mIoU at 80 epochs), while achieving accuracy comparable to a much
   longer-trained MoCo.%
}
\label{table:finetune_result}
\end{table}

Training with a similar siamese architecture on PASCAL+COCO, our method
outperforms InstFeat~\cite{ye2019unsupervised} by a large margin
(46.51~\vs~38.11 mIoU).  We outperform SegSort~\cite{hwang2019segsort}, which
neither leverages information from the region hierarchy nor samples negatives
in a cross-image fashion, by a larger margin (46.51 \vs~36.15).

Though equipped with neither a momentum encoder nor a memory bank, our method
achieves comparable results to a variant of MoCo~\cite{he2019momentum}
(46.51~\vs~47.01 mIoU), while requiring far fewer training epochs (80~\vs~200).
Restricted to only 80 pre-training epochs, we significantly outperform MoCo
(46.51~\vs~42.07).  Given less pre-training data (PASCAL only,
Table~\ref{table:finetune_result} bottom), we similarly observe a substantial
advantage over MoCo (43.51 \vs~32.4 mIoU).  Together, faster convergence and
superior performance with less data suggest that our ability to exploit regions
drastically improves the efficiency of unsupervised visual representation
learning.

This efficiency hypothesis is further supported by the fact that our method
continues to improve if it can take advantage of higher quality edges.  A
variant built using from the advanced edge detector
HED~\cite{xie2015holistically} reaches 48.82 mIoU, outperforming all competing
approaches.  Note, however, that HED itself utilizes ImageNet pre-training, so
this is an ablation-style comparison that does not obey the same strict
restriction, of only relying on unlabeled PASCAL+COCO images, as all other
entries in the corresponding section of Table~\ref{table:finetune_result}.

\begin{figure}[ht!]
   \centering
   \includegraphics[width=1.0\textwidth]{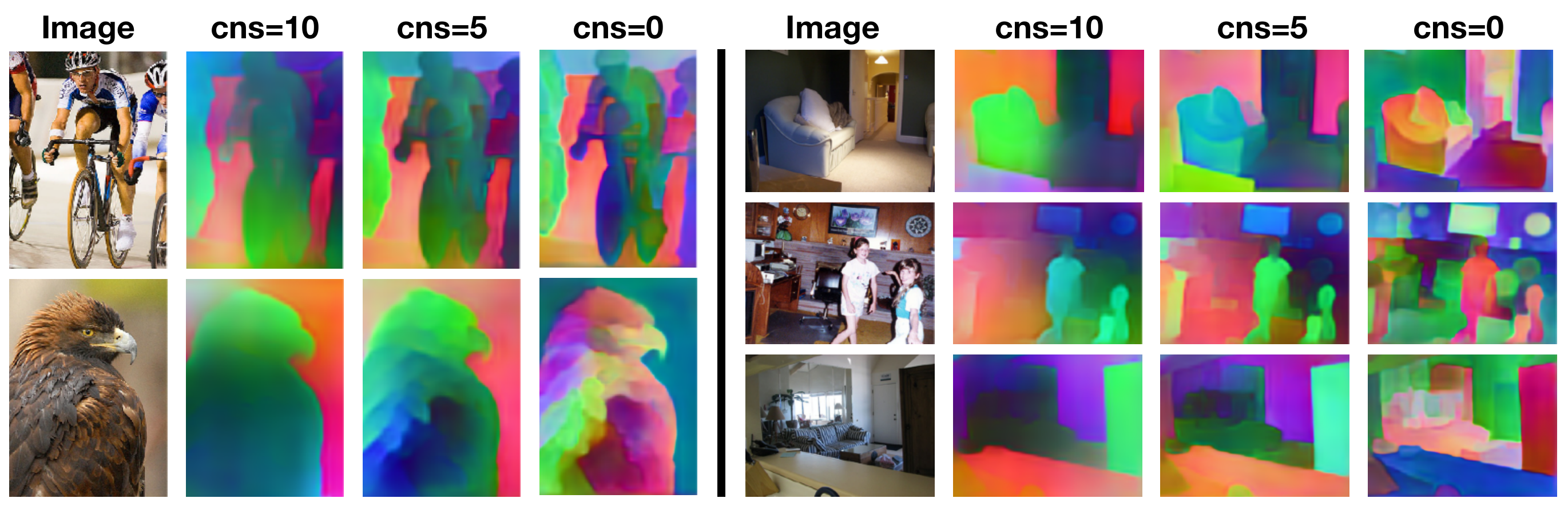}
   \vspace{-15pt}
   \caption{%
      \textbf{Effect of crossing-image negative sampling.}
      Notation $cns=m$  denotes, for each anchor pixel, we randomly sample $m$
      extra features as negative references.  Adjusting $cns$ effectively
      trades-off between learned embeddings that reflect semantic categories
      \vs~local instance boundaries.
   }
   \label{fig:pca_cns}
\end{figure}{}

Besides freezing the backbone, we also run an end-to-end fine-tuning experiment
on a PASCAL+COCO pre-trained backbone, following the evaluation protocol in
He~\etal~\cite{he2019momentum}.  Our method achieves comparable results with
MoCo~\cite{he2019momentum} (47.2 \vs~46.9 mIoU) when both are trained for 80
epochs.  Here, MoCo~\cite{he2019momentum} reaches 55.0 when trained for 200
epochs.

We perform ablations on cross-image negative sampling ($cns$), reported in
Table~\ref{table:finetune_result} (bottom).  Figure~\ref{fig:pca_cns} shows
that increasing $cns$ yields feature embeddings more consistent with semantic
partitioning.

We also experiment with unsupervised training over ImageNet, where MoCo
performs well under the implicit assumption that most unlabeled images only
contain only a single category.  Simply adding our pixel-wise contrastive
target on top of MoCo outperforms the MoCo baseline when benchmarked at 30
epochs (55.13 \vs~53.84), but, as shown in Table~\ref{table:finetune_result}
(top, $\text{Ours}^\dagger$), we do not witness relative improvement at 200
epochs.  However, this does suggest that region awareness can help to boost
unsupervised learning efficiency, even on ImageNet.

\subsection{Semantic Segment Retrieval}

\begin{figure}[t]
   \centering
   \includegraphics[width=\textwidth]{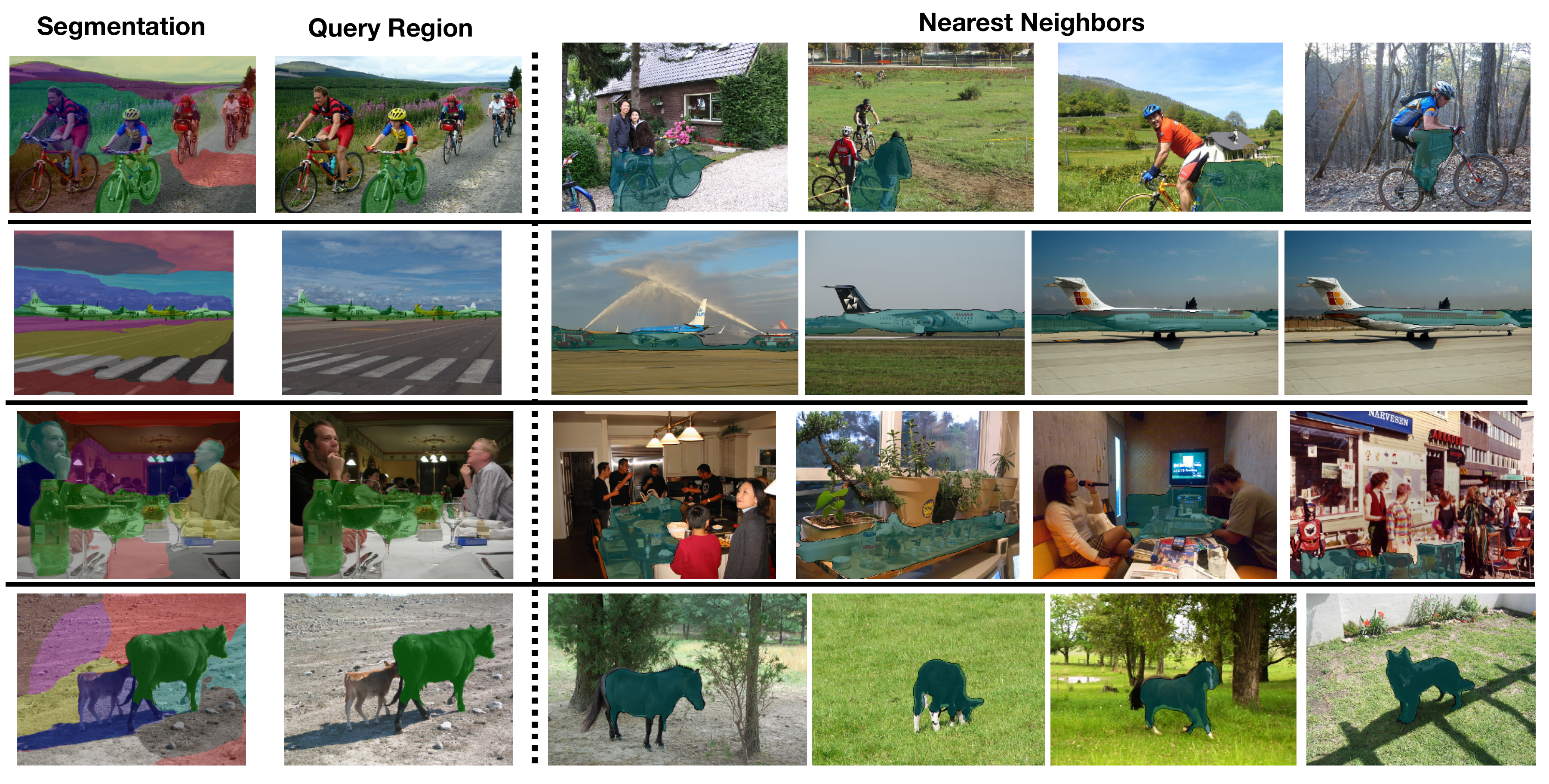}
   \vspace{-15pt}
   \caption{%
      \textbf{Qualitative evaluation of segment search.}
      We leverage our learned embeddings to partition images via K-means
      clustering and serve as region descriptors for nearest neighbor
      search.
   }
   \label{fig:pca_knn}
\end{figure}

\begin{table}[!t]
\centering
\begin{tabular}{c c c c c c c c c }
\toprule
& Mean &Bus & Aero & Car & Person & Cat &  Cow & Bottle \\
\midrule
Class Frequency & - & 4.01\% & 3.00\% & 6.52\% & 25.41\% & 10.66\% & 1.81\% & 1.70\% \\
\midrule
SegSort~\cite{hwang2019segsort} & 10.17 & 9.36 & 22.28 & 11.24 & 30.28 & 17.51 & 0.31 & 0.00\\
\midrule
Ours & \textbf{24.60} & 50.00 & 47.41 & 41.89 & 49.41 & 36.00 & 3.15 & 8.46\\
\bottomrule
\end{tabular}
\vspace{5pt}
\caption{%
   \textbf{Quantitative evaluation of segment search.}
   We report mean and per-class IoU (for selected classes) on PASCAL
   \textit{val}.  Our approach works best on high frequency or rigid object
   classes.
}
\label{table:segment_retrieval}
\end{table}

We also adopt a direct approach to examine our learned embeddings.  We first
partition the images of PASCAL $train\_aug$ and $val$ set into a fixed number of
segments by running K-means (K=15) clustering on the embedding.  Then we use a region
feature descriptor, computed by averaging the feature vectors over all pixels
within a segment, to retrieve nearest neighbors of the $val$ set regions from
the $train\_aug$ set.  We report qualitative (Figure~\ref{fig:pca_knn}) and
quantitative (Table~\ref{table:segment_retrieval}) results.  Without any
supervised fine-tuning, our learned representations reflect semantic categories
and object shape.

\subsection{Instance Mask Tracking}

In this task, we track instance masks by fetching cross-frame neighboring
pixels measured under feature similarity induced by our output embedding.
Specifically, we predict the instance class of pixel $i$ at time step $t$ by
$y_t(i) =
   \sum_{k} \sum_{j} \text{sim}(\bold{F}_t(i),\bold{F}_{t-k}(j))y_{t-k}(j)$,
where $\bold{F}_i(t)$ denotes the feature vector of pixel $i$ at time step $t$,
which we also augmented with spatial coordinates $i$.  We utilize the $k$
previous frames to propagate labels.  $y_t(i)$ is a one-hot vector indicating
the instance class assignment for pixel $i$ at time frame $t$.  In our
experiment, we choose $k = 2$ and $j$ as the set of $10$ nearest neighbors of
pixel $i$.

\begin{figure}[t]
   \centering
   \includegraphics[width=\textwidth]{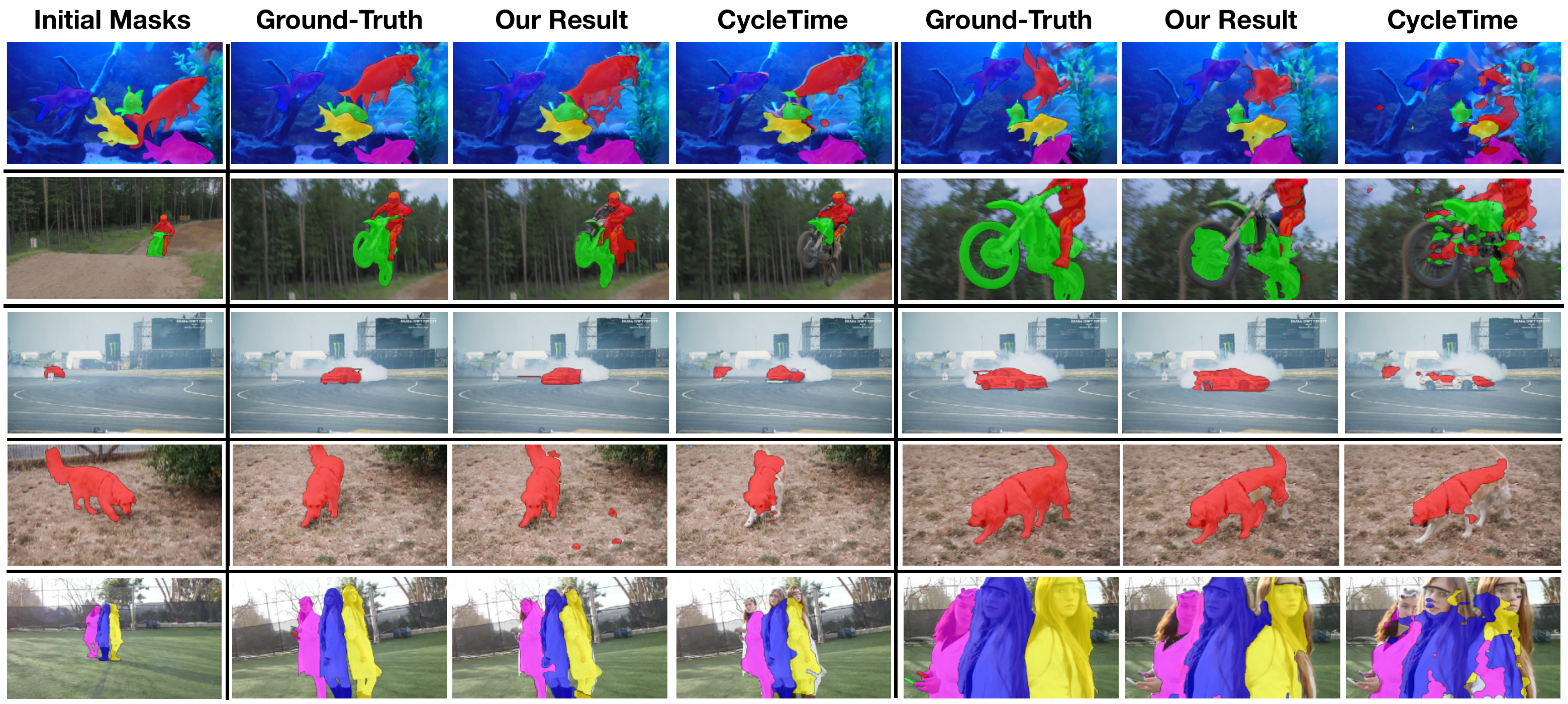}
   \caption{%
      \textbf{Video-based instance tracking on
      DAVIS-2017~\cite{Pont-Tuset_arXiv_2017}.}
      Our method outperforms video-based unsupervised approach
      CycleTime~\cite{wang2019learning} on both region and boundary quality.%
   }
   \label{fig:pca_tracking}
\end{figure}

We evaluate performance using region similarity $\mathcal{J}$ and contour
accuracy $\mathcal{F}$ (as defined by~\cite{perazzi2016benchmark}), with
Table~\ref{table:davis_result} reporting results.  Our feature representation,
learned by respecting the region hierarchy, benefits temporal matching for
difficult cases, \eg~large motion, where local intensity is not reliable.
Though not optimized for precise temporal matching, our approach still
outperforms recent state-of-the-art video-based unsupervised approaches in
both region and boundary quality benchmarks.

\begin{table}[h]
\centering
\begin{tabular}{c@{}cc|ccc@{}}
\toprule

      Method & $\mathcal{J}$(Mean)$\uparrow$ & $\mathcal{F}$(Mean)$\uparrow$ &  Method & $\mathcal{J}$(Mean)$\uparrow$ & $\mathcal{F}$(Mean)$\uparrow$\\
\midrule
     Identity &  22.1 & 23.6& Random Init & 27.7 & 26.6 \\
     FlowNet2 \cite{ilg2017flownet} & 26.7 & 25.2 &
     SIFT Flow \cite{liu2008sift}& 33.0 & 35.0 \\
     DeepCluster \cite{caron2018deep}& 37.5 & 33.2 &
     Video Colorization \cite{vondrick2018tracking} & 34.6 & 32.7 \\
     CycleTime \cite{wang2019learning}& 41.9 & 39.4 &
     mgPFF \cite{kong2019multigrid}& 42.2 & 46.9\\
     \midrule
     Ours ($cns = 0$) & 43.1 & 46.7 &
     Ours ($cns = 5$) & $\mathbf{47.1}$ & $\mathbf{48.9}$\\
\bottomrule
\end{tabular}
\vspace{5pt}
\caption{%
   \textbf{Quantitative evaluation of instance mask tracking on
   DAVIS-2017~\cite{Pont-Tuset_arXiv_2017}.}
   We benchmark region quality ($\mathcal{J}$) and boundary quality
   ($\mathcal{F}$) on the validation set, using the respective metrics defined
   by Perazzi~\etal~\cite{perazzi2016benchmark}.  Our method outperforms
   competitors on both metrics.%
}
\label{table:davis_result}
\end{table}

\section{Conclusion}
We propose a self-supervised learning framework, only leveraging a visual
primitive predictor trained on a small dataset, that bootstraps visual feature
representation learning on large-scale unlabeled image sets by optimizing a
pixel-wise contrastive loss to respect a primitive grouping hierarchy.  We
demonstrate the effectiveness of our pixel-wise learning target as deployed
on unlabeled images of complex scenes with multiple objects, through
fine-tuning to predict semantic segmentation.  We also show that our learned
features can directly benefit the downstream applications of segment search
and instance mask tracking.

\section{Broader Impact}
\label{sec:broader impact}

As an advance in self-supervised visual representation learning, our work
may serve as a technical approach for a wide variety of applications that
learn from unlabeled image datasets, with impact as varied as the potential
applications.  We believe that a compelling and practical use case is likely
to be in domains where human annotation is especially difficult, such as
medical imaging, and are hopeful that that further development of these
techniques will eventually have a positive impact in medical and scientific
domains.

\begin{ack}
The authors have no competing interests.
\end{ack}

\bibliographystyle{splncs04}
\bibliography{egbib}
\end{document}